# Geometric Semantic Genetic Programming Algorithm and Slump Prediction

Juncai Xu*[1], Zhenzhong Shen[1], Qingwen Ren[1], Xin Xie[2], and Zhengyu Yang[2]

*1 College of Water Conservancy and Hydropower Engineering, Hohai University, Nanjing 210098, China*
*2 Department of Electrical and Engineering, Northeastern University, Boston, MA 02115, USA*

ABSTRACT
Research on the performance of recycled concrete as building material in the current world is an important subject. Given the complex composition of recycled concrete, conventional methods for forecasting slump scarcely obtain satisfactory results. Based on theory of nonlinear prediction method, we propose a recycled concrete slump prediction model based on geometric semantic genetic programming (GSGP) and combined it with recycled concrete features. Tests show that the model can accurately predict the recycled concrete slump by using the established prediction model to calculate the recycled concrete slump with different mixing ratios in practical projects and by comparing the predicted values with the experimental values. By comparing the model with several other nonlinear prediction models, we can conclude that GSGP has higher accuracy and reliability than conventional methods.

## 1. Introduction

The rapid development of the construction industry has resulted in a huge demand for concrete, which, in turn, caused overexploitation of natural sand and gravel as well as serious damage to the ecological environment. Such demand produces a large amount of waste concrete in construction, entailing high costs for dealing with these wastes [1-3]. In recent years, various properties of recycled concrete were validated by researchers from all over the world to protect the environment and reduce processing costs. Slump is one of the important indexes reflecting recycled concrete workability and is an important measure of the working performance of recycled concrete. Slump is measured by testing the total wastes of raw materials and manpower. As such, using the intelligent prediction algorithm to directly predict the recycled concrete slump has important practical value [4]. Early researchers predicted concrete slump through the linear regression method. However, determining a satisfactory answer is difficult because of the complex composition of concrete. Consequently, some researchers adopted artificial intelligence, such as neural network [5,6], adaptive fuzzy reasoning methods [7], and genetic programming algorithm [8], to predict concrete slump. The results have shown that these methods can obtain satisfactory results under certain conditions [9,10].

Recently, Moraglio proposed the geometric semantic genetic programming (GSGP) algorithm [11] by introducing the geometric semantic algorithm into the genetic programming algorithm. Given that the size of the retention formula exhibited exponential growth, Vanneschi improved the GSGP, avoided the defect retention formula, and achieved good results in its application to life sciences [12]. Castell applied the GSGP algorithm to predict the strength of high-performance concrete and proved through experiments that it has good reliability [13]. The component of recycled concrete is more complex than that of conventional concrete. As such, how to predict and solve recycled concrete slump using the GSGP algorithm is an important topic.

---

* Corresponding author (E-mail: xujc@hhu.edu.cn)

## 2. Geometric Semantic Genetic Programming Algorithm

Genetic programming algorithm is based on the structuring process through the evolutionary function. This algorithm uses genetic manipulation, including reproduction, crossover, and mutation, to derive the solution of different iterations. The optimal solution will be retained as the result of the problem. The solution of genetic programming can be represented by one tree structure-function expression, which is often regarded as function and terminator sets. However, conventional genetic programming algorithm does not consider the actual meaning of the function. Semantic genetic programming algorithm was developed from conventional genetic programming algorithm. The GSGP algorithm uses the geometric semantic approach instead of the binary tree of the conventional genetic algorithm for crossover and mutation operations.

The specific implementation steps of the GSGP are as follows.

(1) In the initialization process, individuals consist of the function set $F$ and the terminator set $T$ to generate the initial population. The function and terminator sets are expressed as

$$F = \{f_1, f_2, \cdots, f_n\} \quad (1)$$

where $f_i$ denotes the mathematical operation symbols, including $+, -, \times, \div$.

$$T = \{t_1, t_2, \cdots, t_n\} \quad (2)$$

where $t_i$ is the variable comprising the terminator set.

(2) The fitness function is used to evaluate the quality standards for each individual in the population and to calculate the fitness of each individual in the population. This function is also the driving process of evolution to assess each individual. The degree measurement of adaptation methods usually includes original fitness, fitness, and standard normalized fitness. When using the error index, the original fitness can be defined as follows:

$$e(i,t) = \sum_{j=1}^{n} |s(i,j) - c(j)| \quad (3)$$

where $s(i,j)$ is the computational result of individual $i$ in instance $j$, $n$ is the number of instances, and $c(j)$ is the actual result of instance $j$.

(3) Genetic operation consists of parent individual copy and crossover and mutation operations. Crossover operation generates individual $T_C$, and mutation operation generates individual $T_M$ through the parent individual produced using the geometric semantic method, which is expressed as

$$T_c = (T_1 \cdot T_R) + (1 - T_R) \cdot T_2 \quad (4)$$

where $T_1$ and $T_2$ are two parent individuals, and $T_R$ is a real random number.

$$T_M = T + ms \cdot (T_{R1} - T_{R2}) \quad (5)$$

where $T$ is the parent individual, $T_{R1}$ and $T_{R2}$ are two real random functions with codomain in $[0,1]$, and $ms$ is the mutation step.

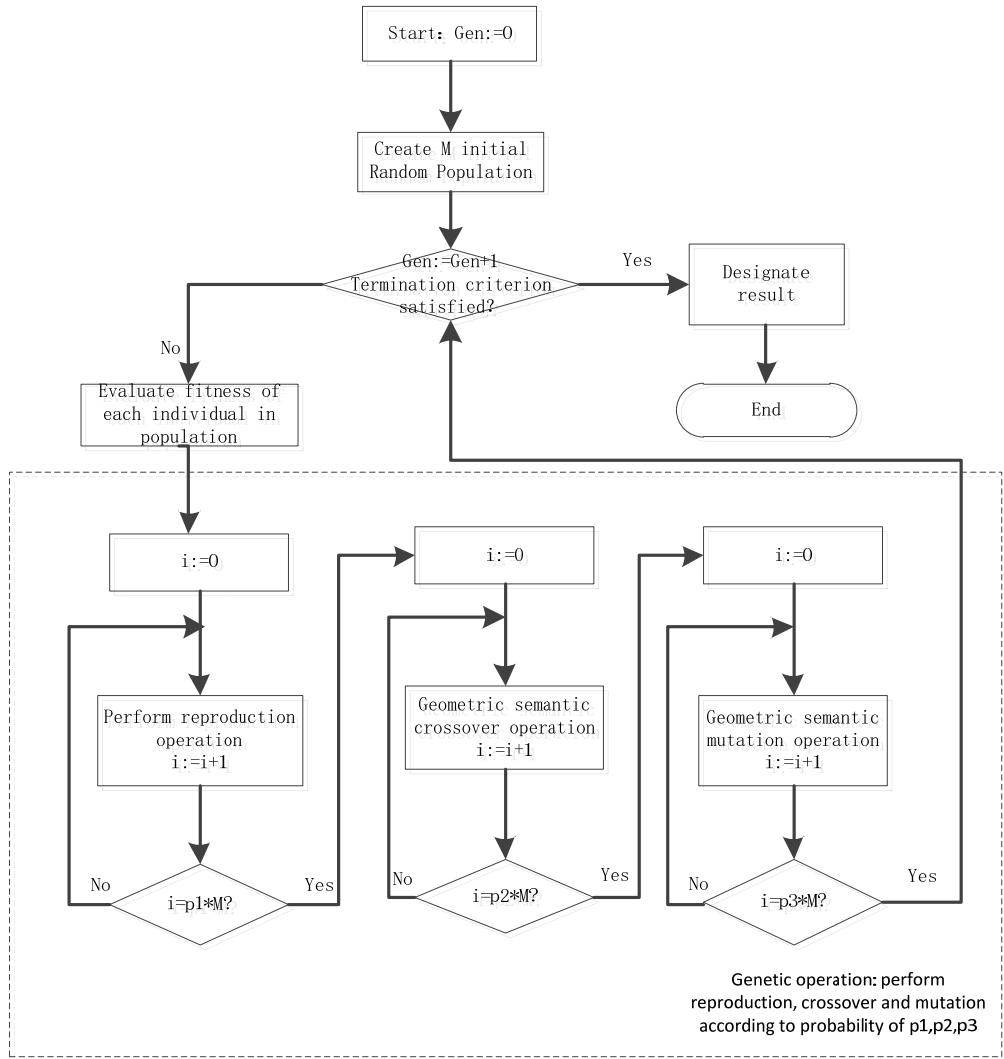

Fig. 1 – Operational structure of geometric semantic genetic programming

Fig. 1 shows the flowchart in the implementation of GSGP on the basis of the aforementioned procedure.

## 3. Genetic programming slump prediction model

For the recycled concrete slump prediction problem, we consider the relevant influence factors based on previous studies and establish the model network structure composed of six input units and one output unit. The input units are factors including cement, fly ash, water, sand, stone, recycled aggregate, water reducer, and total mass. Meanwhile, recycled concrete slump is regarded as the output unit.

A series of evolution parameters of the GSGP has to be determined during the solution process. In the established model, the function set is $\{+,-,\times,\div\}$ and the terminator set has eight variables $\{x_1, x_2, \cdots, x_8\}$ corresponding to cement, fly ash, water, sand, stone, recycled aggregate, water reducer, and total mass. In addition, we set the size of the population, iterative algebraic algorithm, and variation coefficient based on the computational conditions for the genetic programming algorithm. Based on the training samples, the GSGP algorithm can predict the recycled concrete slump for the test samples.

## 4. Engineering application and effect analysis

### 4.1 Engineering of application

Based on experimental data of recycled concrete as dataset (Table 1) [1-4], the established model is applied to predict slump in the research.

Table 1 Concrete mix proportion and experimental slump

| No | cement | fly ash | water | sand | stone | water reducer | recycled aggregate | total mass | slump |
|---|---|---|---|---|---|---|---|---|---|
| 1 | 450 | 0 | 180 | 752 | 1038 | 9.9 | 0 | 2420 | 156 |
| 2 | 400 | 0 | 180 | 769 | 531 | 8.4 | 531 | 2410 | 136 |
| 3 | 317 | 0 | 190 | 787 | 1086 | 5.71 | 0 | 2380 | 125 |
| 4 | 222 | 192 | 185 | 775 | 1070 | 7.4 | 0 | 2400 | 105 |
| 5 | 270 | 234 | 180 | 752 | 1038 | 9.9 | 0 | 2420 | 121 |
| 6 | 333 | 48 | 185 | 775 | 535 | 7.4 | 535 | 2400 | 137 |
| 7 | 254 | 82 | 190 | 787 | 543 | 5.71 | 543 | 2380 | 105 |
| 8 | 333 | 481 | 185 | 775 | 1070 | 7.4 | 0 | 2400 | 150 |
| 9 | 202 | 175 | 185 | 785 | 1084 | 6.38 | 0 | 2390 | 128 |
| 10 | 360 | 117 | 180 | 752 | 1038 | 9.9 | 0 | 2420 | 143 |
| 11 | 240 | 208 | 180 | 769 | 531 | 8.4 | 531 | 2410 | 124 |
| 12 | 400 | 0 | 180 | 769 | 1061 | 8.4 | 0 | 2410 | 149 |
| 13 | 336 | 0 | 185 | 785 | 1084 | 6.38 | 0 | 2390 | 136 |
| 14 | 360 | 117 | 180 | 752 | 519 | 9.9 | 519 | 2420 | 134 |
| 15 | 269 | 87 | 185 | 785 | 1084 | 6.38 | 0 | 2390 | 130 |
| 16 | 202 | 175 | 185 | 785 | 542 | 6.38 | 542 | 2390 | 118 |
| 17 | 296 | 96 | 185 | 775 | 535 | 7.4 | 535 | 2400 | 131 |
| 18 | 370 | 0 | 185 | 775 | 1070 | 7.4 | 0 | 2400 | 150 |
| 19 | 370 | 0 | 185 | 775 | 535 | 7.4 | 535 | 2400 | 138 |
| 20 | 222 | 192 | 185 | 775 | 535 | 7.4 | 535 | 2400 | 120 |
| 21 | 190 | 165 | 190 | 787 | 543 | 5.71 | 543 | 2380 | 105 |
| 22 | 317 | 0 | 190 | 787 | 543 | 5.71 | 543 | 2380 | 108 |
| 23 | 296 | 96 | 185 | 775 | 1070 | 7.4 | 0 | 2400 | 140 |
| 24 | 320 | 104 | 180 | 769 | 1061 | 8.4 | 0 | 2410 | 132 |
| 25 | 259 | 144 | 185 | 775 | 1070 | 7.4 | 0 | 2400 | 120 |
| 26 | 269 | 87 | 185 | 785 | 542 | 6.38 | 542 | 2390 | 120 |
| 27 | 336 | 0 | 185 | 785 | 542 | 6.38 | 542 | 2390 | 126 |
| 28 | 190 | 165 | 190 | 787 | 1086 | 5.71 | 0 | 2380 | 121 |
| 29 | 320 | 104 | 180 | 769 | 531 | 8.4 | 531 | 2410 | 129 |
| 30 | 240 | 208 | 180 | 769 | 1061 | 8.4 | 0 | 2410 | 113 |
| 31 | 259 | 144 | 185 | 775 | 535 | 7.4 | 535 | 2400 | 126 |
| 32 | 450 | 0 | 180 | 752 | 519 | 9.9 | 519 | 2420 | 142 |

| 33 | 270 | 234 | 180 | 752 | 519 | 9.9 | 519 | 2420 | 127 |
| 34 | 254 | 82 | 190 | 787 | 1086 | 5.71 | 0 | 2380 | 123 |

In the process of model testing, the first 28 samples are considered as the training set from the dataset of 34 samples shown in Table 1, whereas the remaining samples are considered the test set. The size of the initial population is set to 500, the maximum iteration epoch is set to 50, and the genetic mutation step is set to 0.1. Table 2 shows the GSGP model prediction and experimental values for slump of the sample. Fig. 2 shows the fitness change with the variation in genetic epoch.

Table 2 Recycled concrete slump using GSGP and experiments.

| No | computation | experiment | No | computation | experiment | No | computation | experiment |
| --- | --- | --- | --- | --- | --- | --- | --- | --- |
| 29 | 130.9 | 129 | 31 | 125.2 | 126 | 33 | 131.9 | 127 |
| 30 | 111.5 | 113 | 32 | 148.6 | 142 | 34 | 128.6 | 123 |

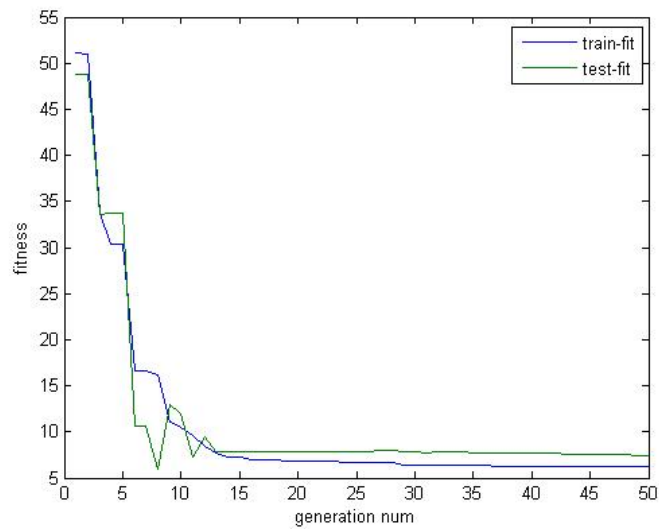

Fig. 2 –  Fitness change with the variation in genetic epoch.

In fitness change with the variation in genetic epoch during the GSGP solving process (Fig. 2), the fitness between the training and testing sets has the same variation rule. The rate of fitness change is sharp before epoch 10. Then, the rate gradually stabilizes with the increase in the number of iterations. The fitness of the test set is slightly larger than that of the training set. The forecast of this method in a smaller genetic algebra is basically stable. Fig. 3 shows the comparison of the GSGP algorithm and multiple linear regression. The prediction of recycled concrete obtained by the test set of the GSGP algorithm is significantly better than that of the multiple linear regression. The relative error of the prediction results is less than 5% from the experiments. As such, recycled concrete slump can be predicted by the GSGP method.

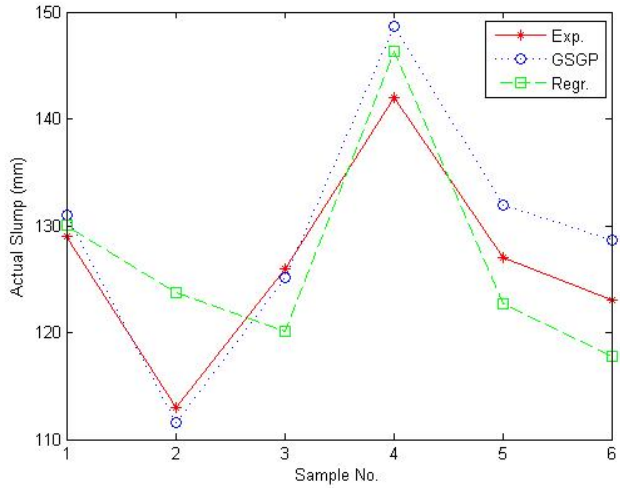

Fig. 3 – GSGP test samples and multiple linear regression predictive value

*4.2 Effect analysis*

We assessed the effects of the GSGP model on predicting recycled concrete slump, which can be reflected by the correlation of the experimental and computational values. The index correlation coefficient between the experimental and computational values is defined as follows:

$$R = \frac{n\sum y \cdot y' - (\sum y)(\sum y')}{\sqrt{\sum y^2 - (\sum y)^2}\sqrt{\sum y'^2 - (\sum y')^2}} \quad (6)$$

where *y* is the experimental value, $y'$ is the computational value, and $n$ is the sample size.

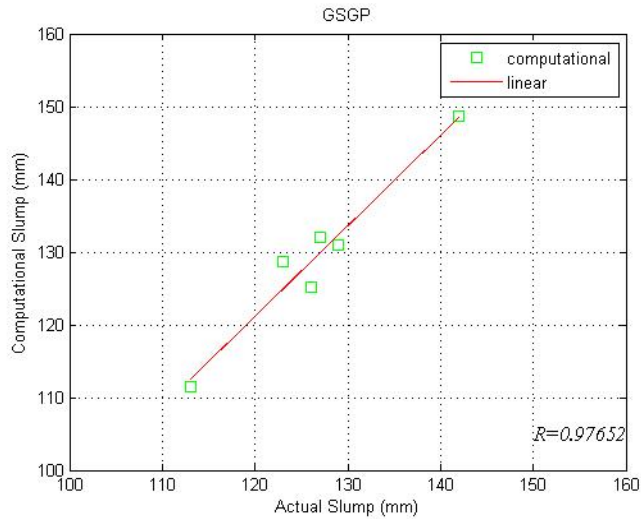

Fig. 4 – Correlation of the experimental and computational values of recycled concrete slump

The correlation coefficient of the experimental and computational values is obtained by using Equation (6). In Fig. 4, the correlation coefficient is greater than 95%. The experimental and computational values have high correlation.

The accuracy of prediction is reflected by the errors. Root mean square error (RMSE) can estimate the error degree, which is defined as follows:

$$RMSE = \sqrt{\frac{\sum(y-y')^2}{n}} \quad (7)$$

where $y$ is the experimental value, $y'$ is the computational value, and $n$ is the sample size.

We further investigated the accuracy of GSGP by comparing its RMSE with two other algorithms, namely, support vector machine (SVM) and standard genetic programming (STGP). The SVM and STPG algorithms were discussed in 15 and 16, respectively. Based on the datasets shown in Table 1, the three algorithms were ran 50 times to predict recycled concrete slump. Fig. 5 shows the statistical results of the RMSE of the three algorithms.

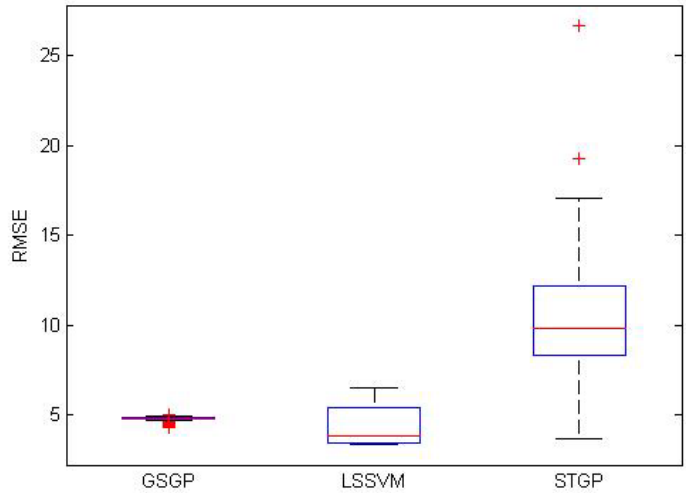

Fig. 5 – Box-whisker plots of the errors of the three algorithms in predicting recycled concrete slump

As shown in Fig. 5, the interquartile ranges of the three algorithms are 0.0892, 1.9933, and 3.8651. The error range using the GSGP algorithm is the narrowest, whereas the error range using the STGP algorithm is the widest. In the Wilcoxon rank–sum analysis, the p of the three algorithms is 1.6387e − 11, 9.5033e − 05, and 2.4373e − 04. The p of the GSGP algorithm is still the lowest of the three algorithms. Thus, the solutions using the GSGP algorithm are significantly better than the other two algorithms.

5. Conclusions

Owing to the complexity of recycled concrete, conventional linear regression cannot obtain satisfactory solutions in predicting recycled concrete slump. The GSGP algorithm was introduced to predict recycled concrete slump. The GSGP model was also established to predict recycled concrete slump, and then applied in concrete engineering. We compared its solution with two other prediction models. The following conclusions were obtained:
(1) The prediction error range is allowable using the GSGP model on the basis of the comparison of the computational and experimental values. Thus, GSGP can be used to predict recycled concrete slump.
(2) The solution of the GSGP model is more accurate than that of the linear regression model. The GSGP model can obtain high-accuracy predicted values of recycled concrete slump.
(3) Based on the slump prediction model of GSGP, the error index is lower than the nonlinear methods, such as SVM and STGP. The algorithm has high suitability and reliability for the prediction of recycled concrete slump.

**Acknowledgments**

This research was funded by the National Natural Science Foundation of China (Grant No. 11132003), A Project Funded by the Priority Academic Program Development of Jiangsu Higher Education Institutions (Grant No.3014-SYS1401).

**References**

Kou S, Poon C, Agrela F. Comparisons of natural and recycled aggregate concretes prepared with the addition of different mineral admixtures. Cement and Concrete Composites, 33(2011)788-795.

Gencel O, Koksal F, Ozel C, Brostow W. Combined effects of fly ash and waste ferrochromium on properties of concrete. Construction and Building Materials, 29(2012) 633-640.

Duan ZH, Poon CS. Properties of recycled aggregate concrete made with recycled aggregates with different amounts of old adhered mortars. Materials & Design, 58 (2014) 19-29.

Yeh I. Modeling slump of concrete with fly ash and superplasticizer. Computers and Concrete , 5(2008) 559-572.

Yeh I. Modeling slump flow of concrete using second-order regressions and artificial neural networks. Cement and Concrete Composites, 29 (2007) 474-480.

Duan Z, Kou S, Poon C. Using artificial neural networks for predicting the elastic modulus of recycled aggregate concrete. Construction and Building Materials, 44 (2013)524-532.

Sobhani J, Najimi M, Pourkhorshidi AR, Parhizkar T. Prediction of the compressive strength of no-slump concrete: a comparative study of regression, neural network and ANFIS models. Construction and Building Materials 24 (2010)709-718.

Chandwani V, Agrawal V, Nagar R. Modeling slump of ready mix concrete using genetic algorithms assisted training of Artificial Neural Networks. Expert Systems with Applications 42(2015) 885-893.

Deshpande N, Londhe S, Kulkarni S. Modeling compressive strength of recycled aggregate concrete by Artificial Neural Network, Model Tree and Non-linear Regression, (2014)187-198.

Duan Z, Poon C. Factors affecting the properties of recycled concrete by using neural networks. COMPUTERS AND CONCRETE, 14 ( 2014) 547-561.

Moraglio A, Krawiec K, Johnson CG. Geometric semantic genetic programming Parallel Problem Solving from Nature-PPSN XII: Springer, (2012)21-31.

Vanneschi L. Improving genetic programming for the prediction of pharmacokinetic parameters. Memetic Computing, 6 (2014) 255-262.

Castelli, M. and Vanneschi, L. et al. Prediction of high performance concrete strength using genetic programming with geometric semantic genetic operators, Expert Systems with Applications, 40 (2013) 6856-6862.

Yali S, Xiaohui L, Yan Li. Prediction about recycled concrete carbonation slump. Concrete , 7(2013)81-78.

Pelckmans K, Suykens JA, Van Gestel T, De Brabanter J, Lukas L, Hamers B, De Moor B. L S-SVMlab: a matlab/c toolbox for least squares support vector machines (2002).

Silva S, Almeida J. GPLAB-a genetic programming toolbox for MATLAB Proceedings of the Nordic MATLAB conference: Citeseer, (2003)273-278.